%% file: main.tex
\documentclass[10pt,twocolumn,letterpaper]{article}

\usepackage{cvpr}              
\usepackage{url}
\usepackage{graphicx}
\usepackage{amsmath}
\usepackage{amssymb}
\usepackage{booktabs}
\usepackage{multirow}
\usepackage{makecell}
\usepackage{animate}
\usepackage[symbol]{footmisc}

\definecolor{cvprblue}{rgb}{0.21,0.49,0.74}
\usepackage[pagebackref,breaklinks,colorlinks,allcolors=cvprblue]{hyperref}

\def\paperID{1958} 
\def\confName{CVPR}
\def\confYear{2025}

\title{\modelname: Continual Text-to-Video Pre-training with Model Expansion and Language Understanding Enhancement}

\author{
    Zhefan Rao\textsuperscript{\rm 1}\footnotemark[1]\ , 
    Liya Ji\textsuperscript{\rm 1}\footnotemark[1]\ , 
    Yazhou Xing\textsuperscript{\rm 1},
    Runtao Liu\textsuperscript{\rm 1},
    Zhaoyang Liu\textsuperscript{\rm 1},
    Jiaxin Xie\textsuperscript{\rm 1}, \\
    Ziqiao Peng\textsuperscript{\rm 2},
    Yingqing He\textsuperscript{\rm 1}\footnotemark[2] ,
    Qifeng Chen\textsuperscript{\rm 1}\footnotemark[2] \\
    \textsuperscript{\rm 1} Hong Kong University of Science and Technology \\
    \textsuperscript{\rm 2}Renmin University of China \\
}

\newcommand{\modelname}{ModelGrow\xspace}

\begin{document}
\input{figure/teaser}
\input{section/abstract}
\footnotetext[1]{These authors contributed equally.}
\footnotetext[2]{Corresponding authors}
\footnotetext{Project webpage: {\color{blue}https://modelgrow.github.io/}}
\input{section/introduction}
\input{section/related_work}
\input{section/method}
\input{section/experiment}
\input{section/conclusion}

\input{section/sup}

{
    \small
    \bibliographystyle{ieeenat_fullname}
    \bibliography{main}
}

\end{document}

%% file: figure/teaser.tex
\twocolumn[{
\maketitle
\begin{center}
    \captionsetup{type=figure}
%
\begin{subfigure}[b]{1\textwidth}
    \hspace{0mm}\includegraphics[width=1\linewidth]{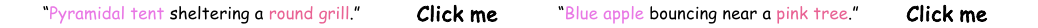}
\end{subfigure}

\setcounter{figure}{0}
\begin{tabular}{c@{}@{}c@{\hspace{1mm}}c@{\hspace{1mm}}c@{\hspace{1mm}}c@{\hspace{1mm}}c@{\hspace{1mm}}c@{\hspace{1mm}}c@{\hspace{1mm}}c@{\hspace{1mm}}}
    \centering

    \put(-8,4){\rotatebox{90}{\small{\textbf{Base model}}}} &
    \includegraphics[width=0.11\linewidth]{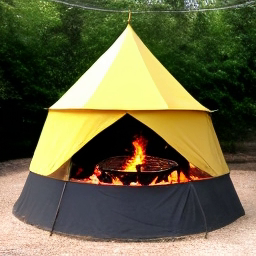}\hspace{-2pt}&
    \includegraphics[width=0.11\linewidth]{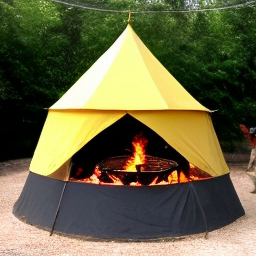}\hspace{-2pt}&
    \includegraphics[width=0.11\linewidth]{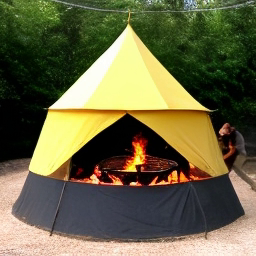}&
    \vrule \hspace{0.5pt}
    \animategraphics[controls=none, width=0.11\textwidth]{8}{asset/teaser/stage4/4/video/}{0}{15}&
    \includegraphics[width=0.11\linewidth]{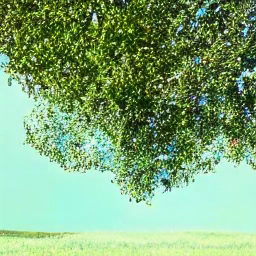}\hspace{-2pt}&
    \includegraphics[width=0.11\linewidth]{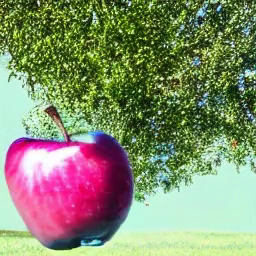}\hspace{-2pt}&
    \includegraphics[width=0.11\linewidth]{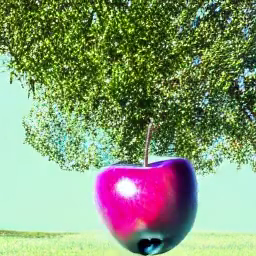}&
    \vrule \hspace{0.5pt}
    \animategraphics[controls=none, width=0.11\textwidth]{8}{asset/teaser/stage4/5/video/}{0}{15}\\

    \put(-8,16){\rotatebox{90}{\small{\textbf{Ours}}}} &
    \includegraphics[width=0.11\linewidth]{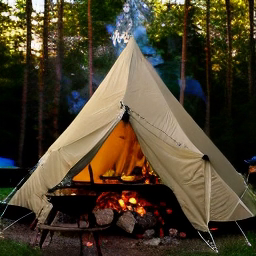}\hspace{-2pt}&
    \includegraphics[width=0.11\linewidth]{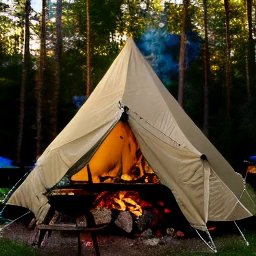}\hspace{-2pt}&
    \includegraphics[width=0.11\linewidth]{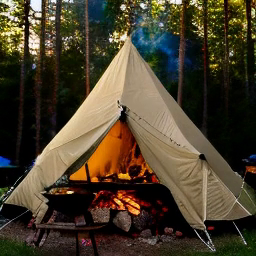}&
    \vrule \hspace{0.5pt}
    \animategraphics[controls=none, width=0.11\textwidth]{8}{asset/teaser/lc_p2_llm/4/video/}{0}{15}&
    \includegraphics[width=0.11\linewidth]{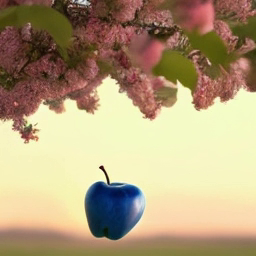}\hspace{-2pt}&
    \includegraphics[width=0.11\linewidth]{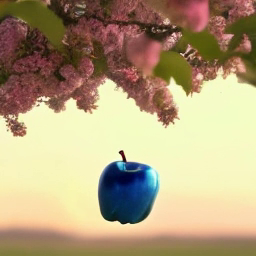}\hspace{-2pt}&
    \includegraphics[width=0.11\linewidth]{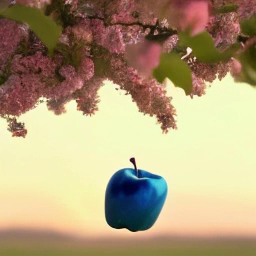}&
    \vrule \hspace{0.5pt}
    \animategraphics[controls=none, width=0.11\textwidth]{8}{asset/teaser/lc_p2_llm/5/video/}{0}{15}\\
\end{tabular}

\captionof{figure}{We continue the pre-training of a text-to-video diffusion model with \modelname, which includes model expansion and language understanding enhancement. 
Our proposed \modelname enhances visual quality, content richness, motion quality, and the ability to prompt following.
For each example, we present three keyframes along with a video. 
Play the video by clicking it with Adobe Acrobat.
}
\end{center}
}]

%% file: section/abstract.tex
\begin{abstract}

Text-to-video (T2V) generation has gained significant attention recently. 
However, the costs of training a T2V model from scratch remain persistently high, and there is considerable room for improving the generation performance, especially under limited computation resources. 
This work explores the continual general pre-training of text-to-video models, enabling the model to ``grow" its abilities based on a pre-trained foundation, analogous to how humans acquire new knowledge based on past experiences.
There is a lack of extensive study of the continual pre-training techniques in T2V generation.
%
In this work, we take the initial step toward exploring this task systematically and propose \modelname.
Specifically, we break this task into two key aspects: increasing model capacity and improving semantic understanding. 
For model capacity, we introduce several novel techniques to expand the model size, enabling it to store new knowledge and improve generation performance. 
For semantic understanding, we propose a method that leverages large language models as advanced text encoders, integrating them into T2V models to enhance language comprehension and guide generation results according to detailed prompts. 
This approach enables the model to achieve better semantic alignment, particularly in response to complex user prompts.
Extensive experiments demonstrate the effectiveness of our method across various metrics. The source code and the model of \modelname will be publicly available.

\end{abstract}




%% file: section/introduction.tex
\section{Introduction}
Investigating general methods for continual training of a text-to-video pre-trained model is gaining more interest in generative areas.
Continual general pre-training aims to enable the model to acquire knowledge from new data that is similar in the domain to the original training data while retaining the information previously learned~\cite{ke2023continual}.
Compared to re-training, this task benefits us by conserving significant computational resources and enhancing the foundation model's general performance by leveraging available pre-trained models in the T2V community.
Existing research~\cite{wu2024llama, ke2023continual} in natural language processing has indicated the success of continual pre-training, which has not been explored in T2V generation yet. In this paper, we make the first step towards continual T2V pre-training instead of achieving state-of-the-art performance.

Continual pre-training for text-to-video generation models faces two challenges: catastrophic forgetting~\cite{de2021continual} and the need for enhanced language understanding when dealing with long and detailed prompts.
First, directly fine-tuning the model on the customized dataset would lead to the model performance drop in the general domain due to catastrophic forgetting.
Especially for text-to-video generation, which requires lots of computational resources and training data, we usually need to train the foundation models with multiple phases.
Second, due to recaptioning~\cite{betker2023improving} being commonly used in the text-to-video generation, the long prompts generated by large language models (LLMs) bring different distributions and higher requirements to generative models, leading to decreased performance of the semantics consistency, both spatially or temporally.
Only utilizing the CLIP~\cite{radford2021learning} or T5~\cite{raffel2020exploring} as the text encoder of diffusion models is not enough to understand the long prompt with detailed information.
It is still an under-explored area of how to enhance language understanding with large language models under continual pretraining.

Expanding the parameters of models could alleviate the forgetting of knowledge and increase the generation ability. 
LLaMA Pro~\cite{wu2024llama}, a block expansion method with only updates the new parameters, proves its effectiveness in natural language generation areas. 
Our work differs from LLaMA Pro in that we focus on the task of text-to-video generation, and we aim to provide practical and insightful guidelines for model expansion with extensive study.
In the direction of LLMs enhancement, current work~\cite{hong2023direct2v, lian2023llm} only focuses on a zero-shot method that acts as a prompt planner to increase the generation ability with LLMs.
This trend will not change the denoising network architecture and thus can not increase the ability of language understanding of the generation model.
We are the first work that systematically explores enhancing the text-to-video generation model both on prompt refining and language understanding improvements via incorporating the embeddings of LLMs.

We propose two continual pre-training methods for text-to-video generation, including model expansion and LLMs enhancement.
For better utilization of the given pretrained model, we introduce an expansion of the transformer block to effectively incorporate new knowledge while minimizing the potential for forgetting previously learned information.
The parameters of the new blocks are duplicated from adjacent blocks to ensure smooth and efficient training. 
This approach not only facilitates the incorporation of additional data but also maintains the structural integrity and performance consistency of the existing model.
Furthermore, we first modify the current re-captioning method for more rich contents and incorporate the embeddings of LLMs as an extra condition in the architecture of the Diffusion Transformer.
Specifically, we expand the cross-attention block for the text condition and duplicate the weights of the original cross-attention block as the initialization of the new block conditioning on LLMs.
Equipped with the prompt template, another cross-attention block conditioning on LLMs embeddings significantly increases the language understanding ability and produces videos with more semantic consistency and vivid motion.

We conduct continual pre-training on our dataset with long prompts generated by LLaVA-NeXT~\cite{li2024llava}.
We also evaluate our models on datasets, VBench~\cite{huang2023vbench} and the subset of CompBench~\cite{sun2024t2v} with quality and semantics metrics.
In summary, our contributions can be summarized as follows:
\begin{itemize}
 \item We propose the model expansion as a continual pre-training method for the transformer-based diffusion model, enabling the incorporation of new knowledge while mitigating the risk of forgetting previously acquired information.
 \item We systematically investigate the LLMs-enhanced text-to-video generation models, both for prompt refining and LLMs embedding incorporation, improving the consistency and quality of generated videos.
 \item 
 We continually train the pre-trained model with our high-quality dataset and conduct extensive evaluations to demonstrate the effectiveness of  \modelname.
\end{itemize}

%% file: section/related_work.tex
\input{figure/method_blocks}
\section{Related Work}
\subsection{Continual Pre-training of Generative Models}
Continual Pre-training~\cite{french1999catastrophic,cossu2024continual} aims to learn from the evolving data without forgetting the knowledge in the past.
In the field of generative models, most of the works~\cite{wu2024llama, ibrahim2024simple, hu2021lora} illustrate the effectiveness of the continual learning strategy in Language Model applications.
The first direction tries to explore the parameter-efficient approaches, adopting the pre-trained model to specific domains, like LoRA~\cite{hu2021lora}, and Adapters~\cite{houlsby2019parameter}.
Another direction in Language Models~\cite{wu2024llama,ibrahim2024simple,wang2023trace} is to continue learning the pre-trained model to enhance the overall performance without decreasing the training scale.
\cite{wu2024llama} introduce a block expansion strategy and only update the new parameters.
\cite{ibrahim2024simple} propose a simple and scalable learning rate strategy and update all the parameters to improve the overall performance.
Unlike post-training, which involves fine-tuning, alignment, and evaluation stages, in this paper, we mainly focus on the general continual pre-training techniques to enhance the quality of generation and language understanding.
To the best of our knowledge, we are the first comprehensive study of continual pre-training in text-to-video generation.

\subsection{Text-to-Video Generation}
Text-to-video generation challenges us to convert low-dimensional data, such as short text, to high-dimensional modality video.
Initially, most of the video generation models are GAN/VAE-based~\cite{van2017neural,mittal2017sync,li2018video,deng2019irc}, which generate videos by learning latent representations and producing sequences of frames that mimic real-world video data.
Recently, the video generation models are based on the diffusion model. We can categorize them into two parts: U-Net-based and transformer-based. The U-Net-based models come first, such as VDM~\cite{ho2022video}, Imagen Video~\cite{ho2022imagen}, LVDM~\cite{he2022latent}, AnimateDiff~\cite{guo2023animatediff}, VideoCrafter1~\cite{chen2023videocrafter1}, VideoCrafter2~\cite{chen2024videocrafter2}, Emu Video~\cite{sun2023emu1,sun2024emu2,wang2024emu3}, WALT~\cite{gupta2025photorealistic}, Lumiere~\cite{BarTal2024LumiereAS}, Show-1~\cite{zhang2023show} and many others~\cite{he-animate-a-story,menapace2024snap,blattmann2023align,ge2023preserve,zhou2022magicvideo,magicvideov2,wang2023lavie}, employ U-Net architectures within diffusion processes to iteratively refine noisy inputs, effectively generating high-quality videos through a series of denoising steps that model complex data distributions. Then, with the OpenAI's Sora~\cite{videoworldsimulators2024} being proposed, more and more transformer-based models appear, such as Open-Sora~\cite{zheng2024opensora}, Open-Sora-Plan~\cite{pku_yuan_lab_and_tuzhan_ai_etc_2024_10948109}, Latte~\cite{ma2024latte},  and CogVideoX~\cite{yang2024cogvideox}, try to utilize the transformer's self-attention mechanisms to capture temporal and spatial dependencies across frames, enabling the generation of coherent and temporally consistent video sequences.

\subsection{Text-to-Video Generation with LLMs}
Incorporating Large Language Models(LLMs) can enhance the generation ability of video foundation models~\cite{zheng2024opensora,he-llm-survey}. 
There are two ways to incorporate LLMs: zero-shot and tuning.
In the zero-shot approach, Re-captioning~\cite{betker2023improving, zheng2024opensora}, a method turning the short prompt into a long-detailed prompt, is commonly used to improve the visual generation quality and accurately follow the user's prompt.
Another direction~\cite{hong2023direct2v, lian2023llm, long2024videodrafter} tries to use LLMs as the planner at the first stage and then generate the whole video given the outputs of the planner.
For tuning, several works~\cite{ge2024seed, yu2023language} aim to produce efficient visual tokens that are suitable for LLMs learning due to the gap between the visual and language modalities.
SEED~\cite{ge2024seed} learns a discrete visual tokenizer by minimizing the reconstruction loss and contrastive loss.
MAGVIT~\cite{yu2023language} proposes a video tokenizer that maps the pixel space into the language domain with Lookup-Free Quantization.
In this paper, we follow the tuning direction but do the opposite, aiming to incorporate the LLMs embeddings into the visual domain to enhance language understanding.

%% file: figure/method_blocks.tex
\begin{figure*}[t]
  \centering
  \includegraphics[width=0.95\linewidth]{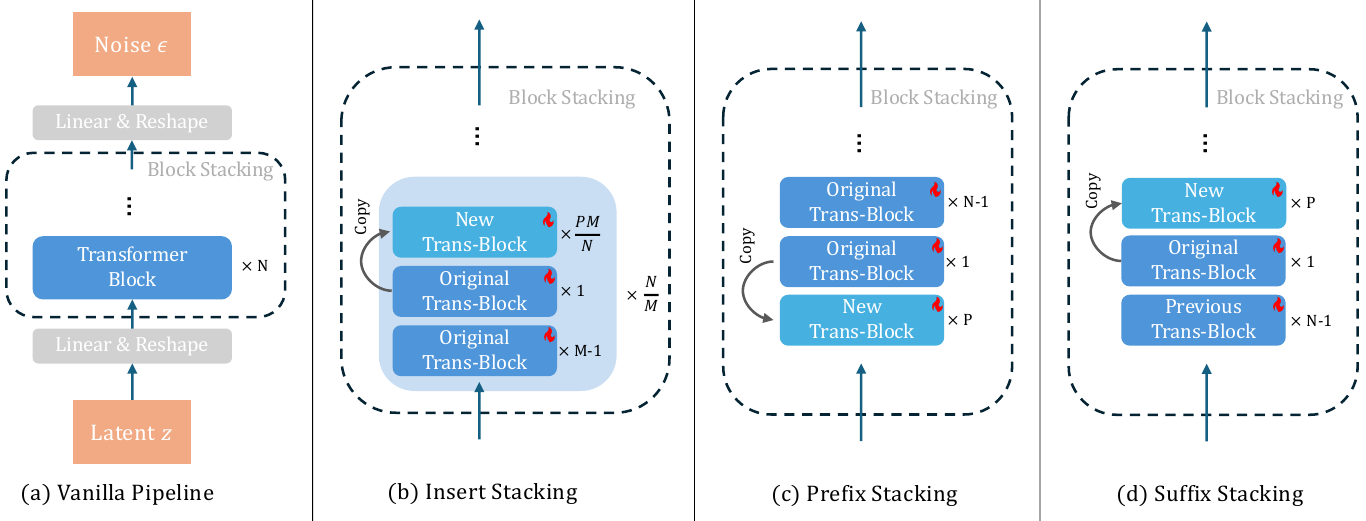}
  \vspace{-2mm}
  \caption{Simplified forward pipeline and variants of block expansion methods. (a) The vanilla pipeline is the traditional stacking of transformer blocks for sequential processing; (b) insert stacking is inserting new transformer blocks intermittently between the existing stack; (c) prefix stacking is adding all the new transformer blocks at the beginning of the stack; (d) suffix stacking is appending all the new transformer blocks at the end of the stack. Each stacking variant illustrates different strategies for arranging transformer blocks to enhance model performance.
  We choose Insert Stacking as our expansion method.}
  \vspace{-5mm}
  \label{fig: blocks}
\end{figure*}

%% file: section/method.tex
\input{figure/method_dit}
\section{Method}
In this section, we introduce the core framework of \modelname. Firstly, we outline the preliminaries that form the basis of our approach in Sec.\ref{method_preliminaries}. Then, we introduce the block expansion process in detail, highlighting its role in enhancing model capacity in Sec.\ref{method_block_expantion}. Finally, we discuss the integration of LLMs in the field of text-to-video generation, demonstrating their impact on performance in Sec.~\ref{method_llm_features}.

\subsection{Preliminary: Diffusion Model with Transformers}\label{method_preliminaries}
\paragraph{Diffusion Architecture}
Diffusion models typically utilize a convolutional U-Net architecture to effectively learn the reverse process necessary for reconstructing the desired output from noise. However, the Diffusion Transformers (DiT)~\cite{peebles2023scalable} had been introduced to replace the U-Net. This transformer is designed to operate on latent patches, thereby enhancing the model's capacity to achieve state-of-the-art performance.

Extending to the video generation model, there are many variants of transformer blocks have been proposed~\cite{ma2024latte}. The base model architecture and the pipeline we choose are shown in Fig.~\ref{fig: blocks} and Fig.~\ref{fig: dit_llm} followed by the Open-Sora V1.0~\cite{zheng2024opensora}. In Fig.~\ref{fig: blocks}, part (a) shows a simplified forward pipeline. We omit the encoding and decoding process of video VAE for the sake of conciseness. The latent features $z \in \mathbb{R}^{B\times C\times T\times H\times W}$ is gotten by the video VAE. It first goes through the linear layer and reshapes the output feature to generate the token embeddings for the next transformer blocks. After several transformer blocks, the embeddings will go through another linear layer and reshaping operation to get the desirable final noise $\epsilon$.

The basic transformer block mainly consists of two self-attention blocks, a cross-attention block, and a feedforward layer. These two sell-attention blocks are the spatial block and the temporal block. The embedding tokens should be reshaped appropriately before going through the corresponding block. After the self-attention blocks, the cross-attention block will absorb the prompt embedding. 

\paragraph{Diffusion Training Loss} To train a diffusion model, we focus on a denoising model $\epsilon_{\theta}$ with parameters $\theta$, which predicts and removes noise at each step. The goal is to minimize the mean squared error between the predicted noise and the actual noise added during the diffusion process:
\begin{equation}
\mathcal{L} = \mathbb{E}_{z\sim p(z),\epsilon\sim\mathcal{N}(0,1)} \left[ \| \epsilon - \epsilon_{\theta}(z_t, t) \|^2 \right]
\label{eq: diffusion_loss}
\end{equation}
In this equation, $z_t$ is the noisy data at step $t$, and $\epsilon$ is the true noise.

where the $w$ denotes the guidance scale.


\subsection{Model Expansion with Block Duplication}\label{method_block_expantion}
\paragraph{Block duplication}
As shown in Fig.~\ref{fig: blocks}, assume the vanilla diffusion model consists of $N$ transformer blocks. 
We propose three block expansion variants according to ~\cite{wu2024llama}. 
Suppose we expand $P$ new transformer blocks, which are copied from the previous blocks. 
Fig.~\ref{fig: blocks} (b) illustrates a variant of block expansion called insert stacking. 
The total block stacking can be divided into $P$ parts. 
Within each part, a new transformer block, duplicated from the previous block, is appended at the end. 
The number of original transformer blocks is $\frac{N}{P}$. Figure~\ref{fig: blocks} (c) depicts another variant of block expansion known as prefix stacking. 
In this approach, all the new transformer blocks are positioned before the original transformer blocks. 
The parameters for these new blocks are duplicated from the first original transformer block. 
Figure~\ref{fig: blocks} (d) presents a variant of block expansion termed suffix stacking. 
In this method, all new transformer blocks are appended after the original transformer blocks. 
The parameters for these additional blocks are duplicated from the last original transformer block.
We conduct the experiments mainly based on the insert stacking variant. 
For our specific configuration, we set $P=N$, meaning that for every original transformer block, a new transformer block is sequentially added immediately following it. 
This configuration allows for a systematic exploration of block expansion effects on model performance. 
Apart from that, we also conduct the ablation study about the different expansion variants and expansion sizes.

\paragraph{Zero initialization}
To maintain outputs following block expansion, we implement a zero initialization procedure for the newly added transformer blocks. 
According to the basic transformer-based diffusion model architecture~\cite{zheng2024opensora}, each transformer block comprises four residual blocks. 
By applying zero initialization, the output of each residual block becomes equivalent to its input, effectively transforming it into an identity block. 
This ensures no alteration in the output initially. 
We achieve this by setting the parameters of the last linear layer within each identity block to zero. 
Specifically, the linear layers within the spatial block, temporal block, cross-attention block, and feed-forward layer are all initialized to zero.

\subsection{Language Understanding Enhancement}\label{method_llm_features}
Given a prompt $p$, we want to generate a video $V$ conditioning on $p$.
Figure~\ref{fig: dit_llm} shows our pipeline of the current LLMs-enhanced transformer block.
To increase the richness of the prompts as well as enhance the understanding ability for the long and complex prompts, we propose two large language model enhancement techniques in text encoder and transformer block. 

\paragraph{Text encoder with LLMs enhancement}
User prompts usually lack details, especially in the text-to-video generation task.
Inspired by the re-captioning in DALLE$\cdot$3~\cite{betker2023improving} and Sora~\cite{videoworldsimulators2024}, we first use LLMs to generate the detailed prompts $p_l$ from the prompt $p$.
Different from the current work that replaces $p$ with $p_l$, we generate $p_{sl}$ by appending $p$ with $p_l$ so that we can highlight the key information as well as maintain the details of the description of the key points.
Before feeding the prompt into the transformer block, except for the original text encoder $\xi(\cdot)$, such as T5~\cite{raffel2020exploring}, for getting the embeddings of the text, we add another text encoder $\xi^*(\cdot)$, reasoned with LLMs parallelly.
Specifically, in order to fit in the distribution of Large Language Models training datasets, we equip $p_{sl}$ with an LLMs template and thus generate $p^*$. An example of an LLMs template from Llama3 is shown in Figure~\ref{fig: dit_llm}.

\input{figure/main_result}

\input{table/main_table}
\paragraph{Transformer block with LLMs enhancement}
We also modify the architecture of the transformer block to enhance the language understanding ability of the video generation model.
We add an extra cross-attention module conditioning on LLMs embeddings following the T5 cross-attention module in every transformer block.
Like the T5 text embedder, we also learn an LLMs-based embedder to adapt to the change of feature distributions.
Inspired by Lumina-T2X~\cite{gao2024lumin-t2x}, we apply gate parameter $\lambda$ followed by the tanh function, leading to a zero-initialization setting.
The text encoder $\xi(\cdot)$ and $\xi^*(\cdot)$ will be frozen, and we will update all the parameters in the transformer block.
Similar to model expansion, we copy the weights of the T5 cross-attention block as the initialization of the weights of the LLMs cross-attention block.
\paragraph{Training loss}
We regard T5 and LLMs embeddings as two separate conditions for video generation.
This setting could allow us flexibly to adjust the influence of LLMs enhancement.
In our text-to-video generation network, the denoising network $\epsilon_\theta(z_t, t, c_{t5}, c_{llm})$ has two conditions, which denotes the T5 embeddings  $c_{t5}$ and the LLMs embeddings $c_{llm}$ separately.
According to the original diffusion training loss Eq.~\ref{eq: diffusion_loss}, our training loss is formulated as follow:
\begin{equation}
\mathcal{L} = \mathbb{E}_{z,t,c_{t5},c_{llm},\epsilon\sim\mathcal{N}(0,1)} \left[ \| \epsilon - \epsilon_{\theta}(z_t, t, c_{t5}, c_{llm}) \|^2 \right].
\label{eq: ours_loss}
\end{equation}
\paragraph{Classifier-free guidance for two conditions} Practically, during the training process, we randomly drop the LLMs condition $c_{llm}$ at $1\%$ and drop all two conditions at $0.1\%$.
Similar to the two conditions formula in InstructPix2Pix~\cite{brooks2023instructpix2pix}, our two text conditions are:
\begin{equation}
\label{eq: cfg}
\begin{split}
    \Tilde{\epsilon}_\theta( z_t, c_{t5}, c_{llm}) & = \epsilon_\theta(z_t) \\
                            &  + s_{t5} (\epsilon_\theta(z_t, c_{t5}) - \epsilon_\theta(z_t)) \\
                            &  + s_{llm} (\epsilon_\theta(z_t, c_{t5}, c_{llm}) - \epsilon_\theta(z_t, c_{t5})).
\end{split}
\end{equation}
where $s_{t5}, s_{llm}$ denote the guidance scales for T5 embeddings and LLMs embeddings accordingly.

%% file: figure/method_dit.tex


\begin{figure*}[t!]
  \centering
  \includegraphics[width=1\linewidth]{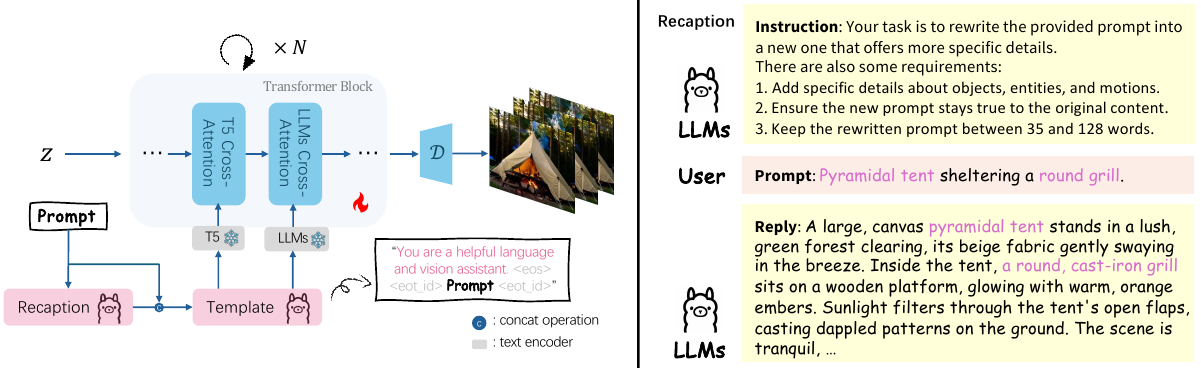}
  \caption{Overview of the pipeline cooperating with the LLMs enhancement. We modify the architecture of the transformer block by adding another cross-attention block, aiming to learn the condition of LLM text embedding. The LLMs cross-attention block follows the original T5 cross-attention block to enhance the language understanding ability of the generation models. For better understanding, we omit the details of the temporal block and spatial block, which are the same as the DiT~\cite{peebles2023scalable} transformer block. All parameters of the transformer block will be updated during the training process.}
  \vspace{-5mm}
  \label{fig: dit_llm}
\end{figure*}

%% file: figure/main_result.tex
\begin{figure*}
    \hspace{3mm}\includegraphics[width=1\linewidth]{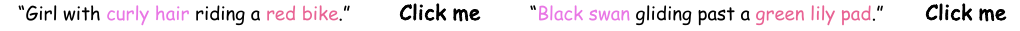}
    \vspace{-8.5mm}
\end{figure*}

\begin{figure*}[t!]
\centering
\begin{tabular}{c@{}@{}c@{\hspace{1mm}}c@{\hspace{1mm}}c@{\hspace{1mm}}c@{\hspace{1mm}}c@{\hspace{1mm}}c@{\hspace{1mm}}c@{\hspace{1mm}}c@{\hspace{1mm}}}
    \centering
    \put(-8,8){\rotatebox{90}{{\footnotesize \textbf{Base model}}}} &
    \includegraphics[width=0.12\linewidth]{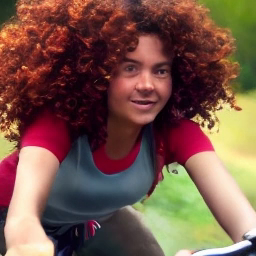}\hspace{-2pt}&
    \includegraphics[width=0.12\linewidth]{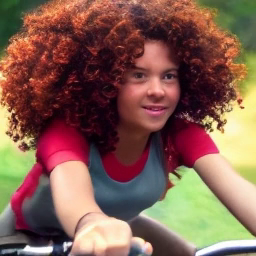}\hspace{-2pt}&
    \includegraphics[width=0.12\linewidth]{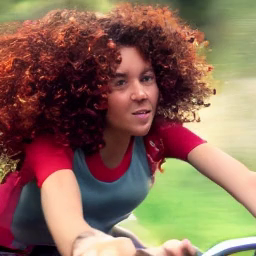}&
    \vrule \hspace{0.5pt}
    \animategraphics[controls=none, width=0.12\textwidth]{8}{asset/figure4/stage4/0/video/}{0}{15}&
    \includegraphics[width=0.12\linewidth]{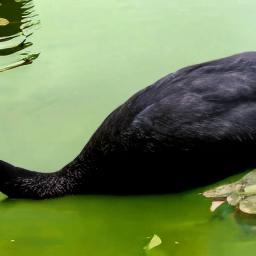}\hspace{-2pt}&
    \includegraphics[width=0.12\linewidth]{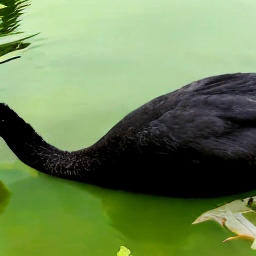}\hspace{-2pt}&
    \includegraphics[width=0.12\linewidth]{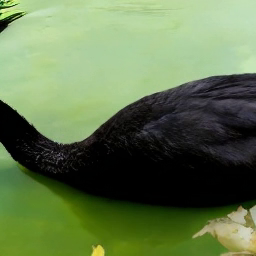}&
    \vrule \hspace{0.5pt}
    \animategraphics[controls=none, width=0.12\textwidth]{8}{asset/figure4/stage4/6/video/}{0}{15}\\

    \put(-8,8){\rotatebox{90}{{\footnotesize \textbf{LoRA-0.7B}}}} &
    \includegraphics[width=0.12\linewidth]{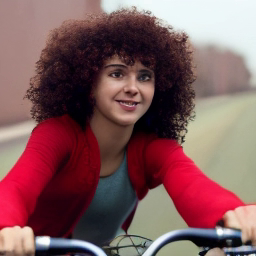}\hspace{-2pt}&
    \includegraphics[width=0.12\linewidth]{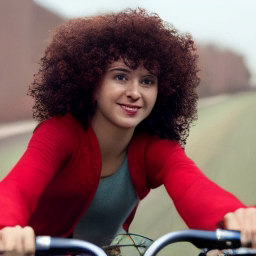}\hspace{-2pt}&
    \includegraphics[width=0.12\linewidth]{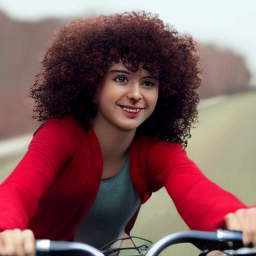}&
    \vrule \hspace{0.5pt}
    \animategraphics[controls=none, width=0.12\textwidth]{8}{asset/figure4/lc_p_lora/0/video/}{0}{15}&
    \includegraphics[width=0.12\linewidth]{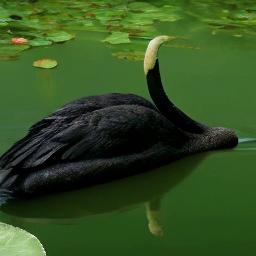}\hspace{-2pt}&
    \includegraphics[width=0.12\linewidth]{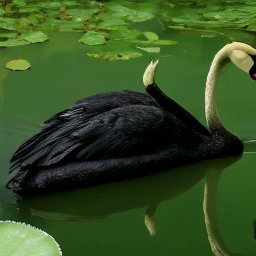}\hspace{-2pt}&
    \includegraphics[width=0.12\linewidth]{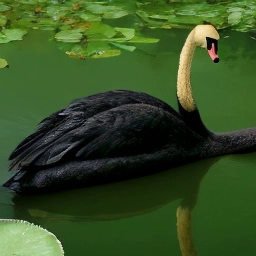}&
    \vrule \hspace{0.5pt}
    \animategraphics[controls=none, width=0.12\textwidth]{8}{asset/figure4/lc_p_lora/6/video/}{0}{15}\\
    
    \put(-8,4){\rotatebox{90}{{\footnotesize \textbf{Expansion-1.4B}}}} &
    \includegraphics[width=0.12\linewidth]{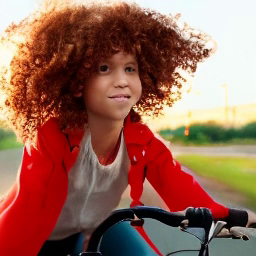}\hspace{-2pt}&
    \includegraphics[width=0.12\linewidth]{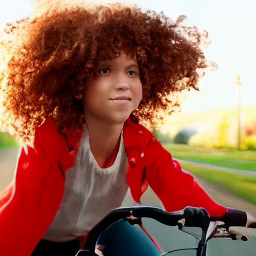}\hspace{-2pt}&
    \includegraphics[width=0.12\linewidth]{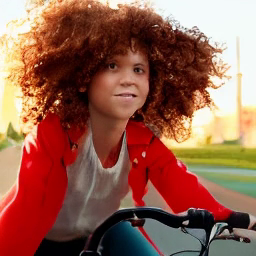}&
    \vrule \hspace{0.5pt}
    \animategraphics[controls=none, width=0.12\textwidth]{8}{asset/figure4/lc_p2/0/video/}{0}{15}&
    \includegraphics[width=0.12\linewidth]{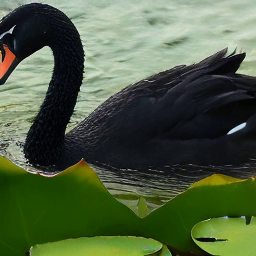}\hspace{-2pt}&
    \includegraphics[width=0.12\linewidth]{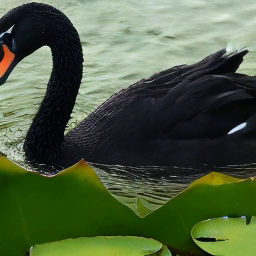}\hspace{-2pt}&
    \includegraphics[width=0.12\linewidth]{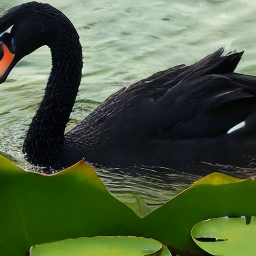}&
    \vrule \hspace{0.5pt}
    \animategraphics[controls=none, width=0.12\textwidth]{8}{asset/figure4/lc_p2/6/video/}{0}{15}\\
    
    \put(-8,-2){\rotatebox{90}{{\scriptsize \textbf{Expansion-1.4B-LLM}}}}
    &
    \includegraphics[width=0.12\linewidth]{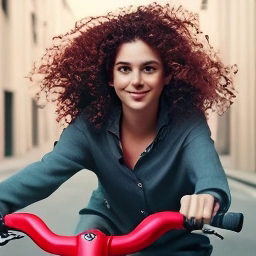}\hspace{-2pt}&
    \includegraphics[width=0.12\linewidth]{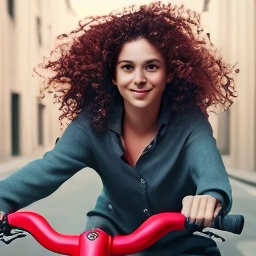}\hspace{-2pt}&
    \includegraphics[width=0.12\linewidth]{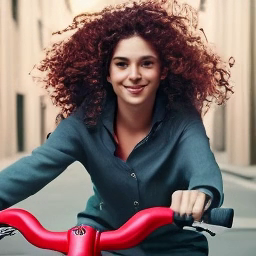}&
    \vrule \hspace{0.5pt}
    \animategraphics[controls=none, width=0.12\textwidth]{8}{asset/figure4/lc_p2_llm_sl/0/video/}{0}{15}&
    \includegraphics[width=0.12\linewidth]{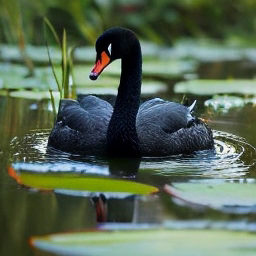}\hspace{-2pt}&
    \includegraphics[width=0.12\linewidth]{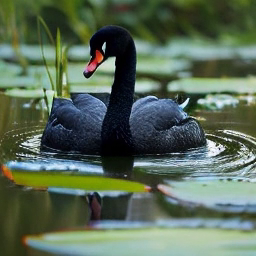}\hspace{-2pt}&
    \includegraphics[width=0.12\linewidth]{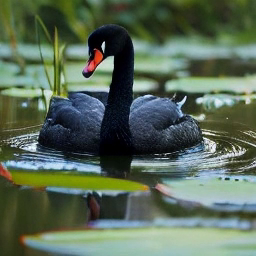}&
    \vrule \hspace{0.5pt}
    \animategraphics[controls=none, width=0.12\textwidth]{8}{asset/figure4/lc_p2_llm_sl/6/video/}{0}{15}\\

\end{tabular}  
\vspace{-2mm}
\captionof{figure}{Qualitative results of our results compared with baselines. Our model Expansion-1.4B-LLM generates videos with higher quality and more semantic alignment than the results of baselines, given the prompt. Play the video by clicking it with Adobe Acrobat.}
\vspace{-2mm}
\label{fig: main_result}
\end{figure*}

%% file: table/main_table.tex
\begin{table*}[ht!]
\centering
\begingroup
\renewcommand{\arraystretch}{1.5} 
\resizebox{\textwidth}{!}{
\setlength{\tabcolsep}{4pt} 
\begin{tabular}{lcccccccc}
    \toprule
    \multicolumn{1}{c}{\multirow{3}{*}{\textbf{Method}}}& \multicolumn{3}{c}{\small{\textbf{VBench Standard Prompts~\cite{huang2023vbench}}}} & \multicolumn{5}{c}{\small{\textbf{T2V-CompBench Consistency Attribute Prompts~\cite{sun2024t2v}}}} \\
    & \makecell[c]{\small{Total}\\ \small{Score}}$\uparrow$ & \makecell[c]{\small{Quality} \\ \small{Score}}$\uparrow$ & \makecell[c]{\small{Semantic}\\ \small{Score}}$\uparrow$ & \makecell[c]{\small{Subject} \\ \small{Consistency}}$\uparrow$ & \makecell[c]{\small{Background} \\ \small{Consistency}}$\uparrow$ & \makecell[c]{\small{Imaging} \\ \small{Quality}}$\uparrow$ & \makecell[c]{\small{Aesthetic} \\ \small{Quality}}$\uparrow$   & \makecell[c]{\small{MLLM} \\ \small{Score}}$\uparrow$  \\ 
    \midrule
    Base model                 & 76.28$\%$          & 78.59$\%$          & 67.04$\%$          & 93.99$\%$          & 95.79$\%$         & 63.78$\%$         & 55.02$\%$        & 0.6015    \\ 
    Vanilla-0.7B                & 76.27$\%$          & 78.73$\%$          & 66.42$\%$          & 95.53$\%$          & 97.01$\%$         & 62.83$\%$         & 56.65$\%$        & 0.6375    \\ 
    LoRA-0.7B~\cite{hu2021lora}   & 77.41$\%$          & 79.86$\%$          & 67.60$\%$          & 96.27$\%$          & 97.07$\%$         & 64.25$\%$         & 57.37$\%$        & 0.6363    \\
    Expansion-1.4B             & 77.98$\%$          & 80.48$\%$ & 67.99$\%$          & 96.50$\%$          & 97.30$\%$         & 66.52$\%$         & 57.33$\%$        & 0.6438    \\ 
    Expansion-1.4B-LLM         & \textbf{78.49$\%$} & \textbf{81.01$\%$}          & \textbf{68.42$\%$} & \textbf{97.00$\%$} & \textbf{97.54$\%$}& \textbf{67.28$\%$}&\textbf{58.46$\%$}& \textbf{0.6737} \\
    \bottomrule
\end{tabular}}
\endgroup
\vspace{-2mm}
\caption{Quantitative results on the VBench standard prompts and CompBench consistency attribute prompts. \textbf{Vanilla} and \textbf{Expansion} denote remaining and expanding the transformer blocks, respectively. \textbf{LLM} denotes introducing the LLMs enhancement. MLLM Score denotes the semantic score reasoned by LLaVa-v1.6-34b~\cite{li2024llava}.}
\vspace{-5mm}
\label{tab: main_reault}
\end{table*}

%% file: section/experiment.tex
\section{Experiments}
In this section, we will introduce the critical experimental information sequentially. Firstly, we outline the foundational experiments' settings in the Sec.~\ref{exp_setup}. Then, we show the main quantitative and qualitative results of our methods in Sec.~\ref{exp_result}. Finally, the ablation study shows the different effectiveness between the various architectures in Sec.~\ref{exp_ablation}.

\input{table/ablation_block}

\subsection{Experimental Setup}\label{exp_setup}
\paragraph{Pretraining stage.}
We utilize a subset of Panda-70M dataset~\cite{chen2024panda70m} including around 2 million video clips, the Vript dataset~\cite{yang2024vript} including around 0.4 million video clips and the Webvid dataset~\cite{bain2021webvid} including around 2 million video clips. Apart from these video datasets, we also use the JourneyDB dataset~\cite{pan2023journeydb}, including around 4 million high-quality images for joint image and video training.

For the base model, we begin with Open-Sora V1.0~\cite{zheng2024opensora}, which is around 0.7 billion parameters, and train the model with the datasets described above to produce our base model.
According to its report, we initialize the base model with the PixArt-$\alpha$~\cite{chen2023pixartalpha}. 
During the training, the batch size is set to 32 on a single GPU, and the learning rate is set to 2e-5.
Each training video clip contains 16 frames with the resolution $256\times 256$. The base model is trained for around 108k steps on 32 NVIDIA H800 GPUs.

\paragraph{Continual pretraining stage.}
In this stage, we collect a new dataset, including 400k high-quality video clips with detailed long prompts generated by the LLaVA-NeXT~\cite{li2024llava}.
An example of our training prompts is illustrated in supplementary materials.
We conduct the training stage on 8 NVIDIA H800 GPUs.
For a fair comparison, we keep the same total number of training steps times total batch sizes. 
As for the training hyper-parameters, we still set the learning rate as 2e-5, the same as the pretrained stage. Each training video clip contains 16 frames with the resolution $256\times 256$.

\paragraph{Evaluation stage.}
We primarily utilize VBench~\cite{huang2023vbench} as our evaluation framework due to its robust and comprehensive suite for assessing video model performance.
The evaluation experiment is structured into two distinct parts. The first part utilizes the standard VBench evaluation suite, which requires that we use the official standard prompts during the inference. This official prompt file contains 946 short prompts.
For the second part, to precisely assess the impact and effectiveness of the LLMs enhancement within our approach, we use the Consistency Attribute Prompts from T2V-CompBench~\cite{sun2024t2v}, including 100 high-quality prompts.
Based on the CompBench prompt, we jointly use the customized VBench evaluation suite and semantic MLLM score reasoned with the LLaVA-v1.6-34b model~\cite{li2024llava} to evaluate the generated results both in quality and semantics.

\input{table/ablation_captionlength}
\input{figure/ablation_recap}
\subsection{Quantitative and Qualitative Results}\label{exp_result}
The main quantitative results are shown in the Table~\ref{tab: main_reault}.
For the VBench standard evaluation, there are three summative scores, which are derived from the analysis of 16 fundamental assessment dimensions.
For the T2V-CompBench prompts, the VBench tool can only evaluate partial dimensions.
The supplementary materials will show the results for all dimensions.
Vanilla-0.7B, LoRA-0.7B, and Expansion-1.4B are trained continually from the Base model, while Expansion-1.4B-LLM are trained from the checkpoints of Expansion-1.4B.
As our baselines, Vanilla-0.7B denotes the vanilla version via updating all parameters from the Base model directly, and LoRA-0.7B represents updating the parameters of the base model in the LoRA~\cite{hu2021lora} approach.

Continually training the pretrained model via model expansion significantly improves the performance compared to vanilla or LoRA training.
The comparison between the Vanilla-0.7B and Expansion-1.4B models reveals that increased model size enhances the model's capability to generate video with higher quality. 
Moreover, when the base model is included in the comparison, it becomes apparent that merely continuing training with the same model size yields limited improvements or even detriments in some dimensions, such as semantic score and imaging quality.
In addition, we also found that we can continually improve the model's performance, especially in semantic alignment, with LLMs enhancement, even though the datasets are still the same.
Table~\ref{tab: main_reault} shows The performance of Expansion-1.4B-LLM improves compared with Expansion-1.4B both in VBench and CompBench Consistency Attribute Prompts in Table~\ref{tab: main_reault}.

We also show some examples perceptually in Figure~\ref{fig: main_result}.
From these examples, we conclude that our model Expansion-1.4B-LLM produces more high-quality generation results and has more semantic alignment with the given prompt. More qualitative results are shown in the supplementary materials.


\subsection{Ablation Study}\label{exp_ablation}
\subsubsection{Ablation Study on Model Expansion}
Table~\ref{tab: ablation_block} presents the quantitative results of an ablation study focusing on model scaling and different variant stacking methods.
Comparing Expansion-2.1B and Expansion-1.1B with Expansion-1.4B shown in Table~\ref{tab: main_reault} reveals the effectiveness of duplicating blocks via increasing model size.
Notably, Expansion-2.1B attains superior results across total, quality, and semantic scores.
Regarding variant stacking, Prefix Stacking enhances the quality score but adversely affects the semantic score.
Conversely, Suffix Stacking improves the semantic score while diminishing the quality score.
Therefore, Insert Stacking is preferred in the main table, as it yields the optimal total score.
\subsubsection{Ablation Study on LLMs Enhancement}
\paragraph{Modified re-captioning} Table~\ref{tab: ablation_length} shows the quantitative results of the ablation study on LLMs Enhancement techniques.
Firstly, we want to verify the effectiveness of modified re-captioning techniques for prompts.
Expansion-1.4B-LLM-S denotes that we infer the Expansion-1.4B-LLM model on the original short prompt.
Expansion-1.4B-LLM-L denotes that we infer the Expansion-1.4B-LLM model on the detailed long prompt reasoned with LLMs.
Expansion-1.4B-LLM-SL denotes that we infer the Expansion-1.4B-LLM model on the concatenation of the original prompt and the detailed long prompt.
The quantitative results in Table~\ref{tab: ablation_length} and the examples in Figure~\ref{fig: ablation_recap} validate the assumption that short prompts maintain the key information, and detailed long prompts can provide more rich content that leads to videos with high quality and semantic consistency.
\input{table/ablation_llm}
\paragraph{Language understanding ability with LLMs injection} We also conduct an ablation study on whether injecting cross-attention block conditioning LLMs embeddings or not, keeping the prompt the same in Table~\ref{tab: ablation_length}.
Comparing Expansion-1.4B-LLM-SL with Expansion-1.4B-SL, we conclude that the model can better understand the prompt with LLMs embeddings and thus generate videos with more semantic alignment. 

\paragraph{The approach incorporating LLMs embeddings} We have proposed two techniques for incorporating LLMs embeddings: equipping the prompt with an LLMs template and duplicating the weights of the T5 cross-attention block to the LLMs cross-attention block as an initialization.
Table~\ref{tab: ablation_llm} shows the quantitative results on VBench Standard Prompts.
We conclude that we can better utilize the LLMs embeddings to enhance the language understanding ability with the two techniques.

\paragraph{Classifier-free guidance} We fix the guidance scale of T5 $s_{t5}$ to 7 and vary the guidance scale of llm $s_{llm}$ from 4 to 20.
Figure~\ref{fig: cfg_curve} shows the quality score and semantic score reasoned by MLLM on the CompBench prompt, respectively.
The larger value $s_{llm}$ indicates the more significant impact of the llm condition on the generated results.
We can see the score has improved with the increase of $s_{llm}$, showing the effectiveness of incorporating LLMs embeddings.
In the paper, we set CFG T5 to 7 and CFG LLMs to 12.5 to run our final model Expansion-1.4B-LLM.
\input{figure/cfg_curve}

%% file: table/ablation_block.tex

\begin{table}[t]
\centering
\resizebox{0.48\textwidth}{!}{
\begingroup
\setlength{\tabcolsep}{5pt} 
\renewcommand{\arraystretch}{1} 
\begin{tabular}{lccc}
\toprule
\multicolumn{1}{c}{\multirow{3}{*}{\textbf{Method}}}& \multicolumn{3}{c}{\textbf{VBench Standard Prompts}}                                               \\
               & Total & Quality  & Semantic    \\
                & Score$\uparrow$ & Score$\uparrow$ &  Score$\uparrow$ \\ \midrule
Expansion-1.1B       & 77.52\%             & 79.94\%               & 67.84\%                          \\ 
Expansion-2.1B       & \textbf{78.37\%}    & \textbf{80.48\%}      & \textbf{69.93\%}                 \\
Expansion-1.4B-Prefix& 77.71\%             & 80.26\%               & 67.50\%                          \\ 
Expansion-1.4B-Suffix& 77.81\%             & 80.00\%               & 69.03\%                          \\
\bottomrule
\end{tabular}
\endgroup}
\vspace{-2mm}
\caption{Quantitative results of ablation study on the different model size and variant stacking methods.
By default, we use Insert Stacking.
\textbf{Prefix} and \textbf{Suffix} denote the model utilizing Prefix Stacking and Suffix Stacking, respectively.}
\vspace{-2mm}
\label{tab: ablation_block}
\end{table}

%% file: table/ablation_captionlength.tex
\begin{table}[t]
\centering
\resizebox{0.48\textwidth}{!}{
\begingroup
\setlength{\tabcolsep}{3pt} 
\renewcommand{\arraystretch}{1} 
\begin{tabular}{lccc}
\toprule
\multicolumn{1}{c}{\multirow{3}{*}{\textbf{Method}}} & 
\multicolumn{3}{c}{\textbf{T2V-CompBench Prompts}}                                               \\
            & Subject & Image & MLLM \\
            & Consistency$\uparrow$ & Quality$\uparrow$ & Score$\uparrow$  \\ \midrule
\multicolumn{4}{l}{LLMs Injection} \\
\midrule
Expansion-1.4B-SL      & 96.18\%              & 65.85\%                & 0.6470                          \\
Expansion-1.4B-LLM-SL  & \textbf{97.00\%}     & \textbf{67.28\%}       & \textbf{0.6737}               \\
\midrule
\multicolumn{4}{l}{Inference Strategy} \\
\midrule
Expansion-1.4B-LLM-S   & 96.52\%             & 66.93\%                & 0.6315                         \\ 
Expansion-1.4B-LLM-L   & 96.73\%           & 65.05\%                & 0.6283                         \\
Expansion-1.4B-LLM-SL  & \textbf{97.00\%}    & \textbf{67.28\%}       & \textbf{0.6737}               \\
\bottomrule
\end{tabular}
\endgroup}
\caption{Quantitative results of ablation study on the method language understanding ability with LLMs injection and modified re-captioning during the inference. The suffix \textbf{S}, \textbf{L}, and \textbf{SL} denote that we infer the model on the original prompt, the detailed long prompt, and the merged prompts of the two types.}
\vspace{-5mm}
\label{tab: ablation_length}
\end{table}

%% file: figure/ablation_recap.tex

\begin{figure}[ht!bp]
\captionsetup[subfigure]{labelformat=empty, justification=justified,singlelinecheck=false}
\begin{subfigure}[b]{0.45\textwidth}
    \hspace{1mm}\includegraphics[width=1\linewidth]{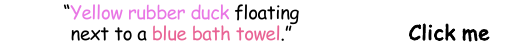}
\end{subfigure}

\setcounter{figure}{4}
\centering
\begin{tabular}{c@{}@{}c@{\hspace{1mm}}c@{\hspace{1mm}}c@{\hspace{1mm}}c@{\hspace{1mm}}}

    \centering
    \put(-8,25){\rotatebox{90}{{\small \textbf{S}}}} &
    \includegraphics[width=0.22\linewidth]{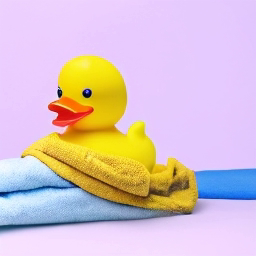}\hspace{-2pt}&
    \includegraphics[width=0.22\linewidth]{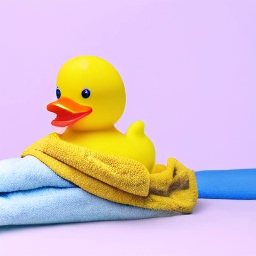}\hspace{-2pt}&
    \includegraphics[width=0.22\linewidth]{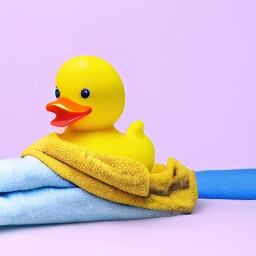}&
    \vrule \hspace{0.5pt}
    \animategraphics[controls=none, width=0.22\linewidth]{8}{asset/length/s/0/video/}{0}{14} \\

    \put(-8,25){\rotatebox{90}{{\small \textbf{L}}}} &
    \includegraphics[width=0.22\linewidth]{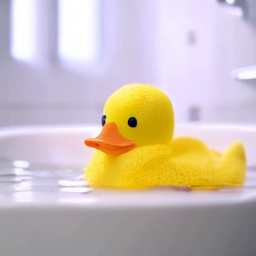}\hspace{-2pt}&
    \includegraphics[width=0.22\linewidth]{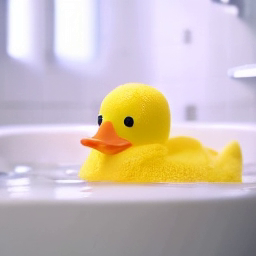}\hspace{-2pt}&
    \includegraphics[width=0.22\linewidth]{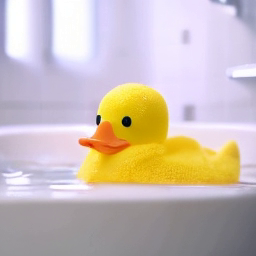}&
    \vrule \hspace{0.5pt}
    \animategraphics[controls=none, width=0.22\linewidth]{8}{asset/length/l/0/video/}{0}{14}\\

    \put(-8,20){\rotatebox{90}{{\small \textbf{SL}}}} &
    \includegraphics[width=0.22\linewidth]{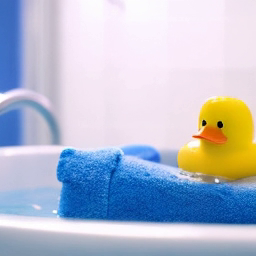}\hspace{-2pt}&
    \includegraphics[width=0.22\linewidth]{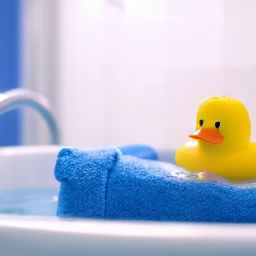}\hspace{-2pt}&
    \includegraphics[width=0.22\linewidth]{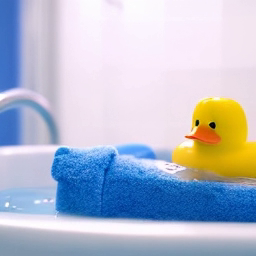}&
    \vrule \hspace{0.5pt}
    \animategraphics[controls=none, width=0.22\linewidth]{8}{asset/length/sl/0/video/}{0}{14}\\
    
\end{tabular}
\begin{subfigure}[b]{0.45\textwidth}
    \includegraphics[width=1\linewidth]{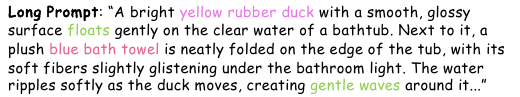}
\end{subfigure}
\captionof{figure}{Examples of our model with different prompts. We use Expansion-1.4B-LLM to evaluate the effectiveness of modified recaptioning. We can see the content of the video is richer with the help of the long prompt (L). However, some key information, such as ``blue bath towel'', is missing if we directly replace the original prompt (S) with the long prompt. By concatenating the original prompt with the long prompt (SL), the model could keep the critical information as well as lead to a richer video result. Play the video by clicking it with Adobe Acrobat.}
\vspace{-6mm}
\label{fig: ablation_recap}
\end{figure}

%% file: table/ablation_llm.tex


\begin{table}[t]
\centering
\resizebox{0.48\textwidth}{!}{
\begingroup
\setlength{\tabcolsep}{2pt} 
\renewcommand{\arraystretch}{1} 
\begin{tabular}{lccc}
\toprule
\multicolumn{1}{c}{\multirow{3}{*}{\textbf{Method}}} & \multicolumn{3}{c}{\textbf{VBench Standard Prompts}}                                               \\
           & Total & Quality  & Semantic    \\
           & Score$\uparrow$ & Score$\uparrow$ &  Score$\uparrow$ \\ \midrule
Vanilla-0.7B-LLM w/o template      & 77.48\%             & \textbf{80.53\%}      & 65.33\%                          \\ 
Vanilla-0.7B-LLM w/o duplication   & 76.01\%             & 80.05\%               & 66.19\%                          \\
Vanilla-0.7B-LLM     & \textbf{77.49\%}    & 80.07\%               &\textbf{67.12\%}                  \\ 

\bottomrule
\end{tabular}
\endgroup}
\vspace{-2mm}
\caption{Ablation study on the approach incorporating LLMs embeddings. 
The pre-trained model is the Base model in Table~\ref{tab: main_reault}.
\textbf{Template} denotes equipping prompt with LLMs template.
\textbf{Duplication} denotes duplicating the weights of the T5 cross-attention block to the LLM cross-attention block as the initialization.}
\vspace{-5mm}
\label{tab: ablation_llm}
\end{table}

%% file: figure/cfg_curve.tex
\setcounter{figure}{5}
\begin{figure}[t]
  \centering
  \includegraphics[width=1\linewidth]{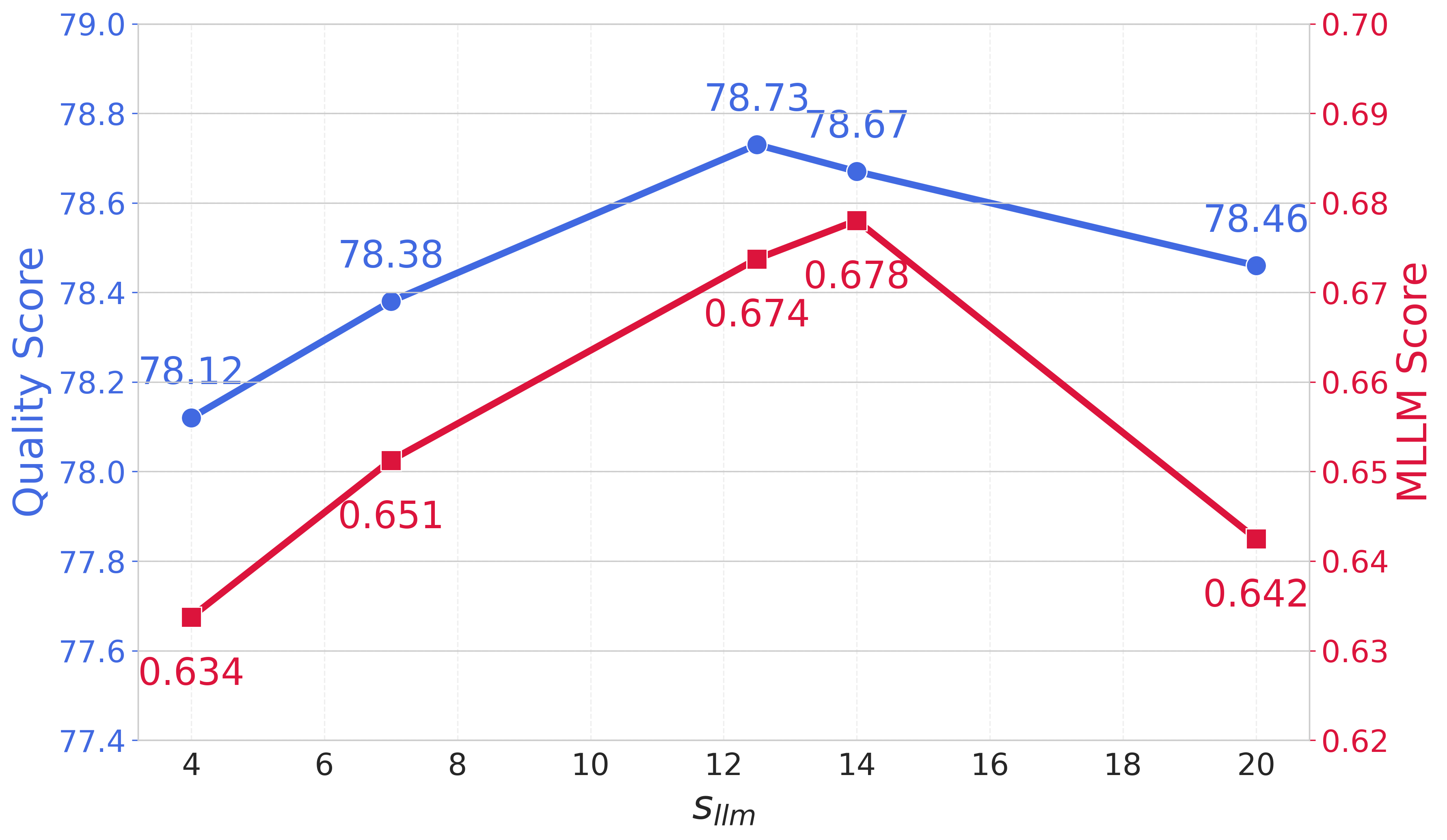}
  \caption{Ablation study on classifier-free guidance (CFG). We fix the scale of CFG T5 to 7 and vary the scale of CFG LLMs from 4 to 20. With the increase of CFG LLM, indicating the influence of the condition of LLMs embeddings, we can improve the generated results of our model.}
  \vspace{-5mm}
  \label{fig: cfg_curve}
\end{figure}

%% file: section/conclusion.tex
\section{Conclusion}
In this paper, we introduced \modelname to tackle the challenges of continual general pre-training in text-to-video generation. The block duplicated expansion technique effectively expands transformer-based diffusion models, allowing them to integrate new knowledge from custom datasets without catastrophic forgetting. By updating all parameters and duplicating blocks from adjacent layers, we ensured smooth integration of new information while preserving the integrity of prior content. Furthermore, we leverage large language models (LLMs) through our LLM-conditioned expansion. By adding LLM embeddings as an extra cross-attention block in the transformer architecture, our model gains better understanding of lengthy and detailed prompts from advanced LLMs.
Due to the high cost of computation resources, we only verify our continual pre-training methods on one base model.
In future work, we would like to apply our methods to other text-to-video foundation models and thus keep improving the performance of video generation ability in the generative community.

%% file: section/sup.tex
\appendix
\section{Details of Insert Stacking}
As shown in Fig.~\ref{supp_fig: stack}, the insert stacking inserts new transformer blocks intermittently between the existing stack. 
Suppose we extend $P=kN$ new transformer blocks, where $k\in \mathbb{R}$ and $P\in \mathbb{N}^+$. 
To express the coefficient $k$ in irreducible fraction form, it can be represented as $\frac{x}{y}$, where $x, y \in \mathbb{N}^+$ and $gcd(x, y) = 1$.
From this, we can get $\frac{PM}{N} = x$ and $M = y$.

To clarify this process, consider the following examples of extending transformer blocks:
\begin{itemize}
    \item In the first case, if we extend $N$ new transformer blocks, where $k=1$. In irreducible fraction form, this is $\frac{1}{1}$. Thus, we can get $M = 1$ and $\frac{PM}{N} = 1$, which means that one new transformer block is placed directly behind each original transformer block in the stack.
    \item In the second case, if we extend $0.5N$ new transformer blocks, where $k=0.5$. Writing this as an irreducible fraction, we have $\frac{1}{2}$. Therefore, we can get $M = 2$ and $\frac{PM}{N} = 1$. Note that $N$ must be divisible by $M = 2$ to maintain divisibility.
    \item In the third case, if we extend $2N$ new transformer blocks, where $k=2$. In irreducible fraction form, this is $\frac{2}{1}$. As a result, we can get $M = 1$ and $\frac{PM}{N} = 2$.
\end{itemize}

\input{supp_asset/figure_stack}
\section{Details of Transformer Block Architecture}
\input{supp_asset/figure_arc}
\input{supp_asset/figure_recap1}
\input{supp_asset/table_vb1}

\input{supp_asset/figure_vs}

We show the detailed architecture of transformer blocks in our work in Fig.~\ref{supp_fig: arc} followed by the Open-Sora V1.0~\cite{zheng2024opensora}. Within each transformer block, the latent feature $z$ is processed through a series of specialized modules: the Spatial Self-Attention Block, Temporal Self-Attention Block, T5 Cross-Attention Block, LLM Cross-Attention Block, and concludes with a Feed Forward Block. This sequence ensures the comprehensive extraction and integration of features across both spatial and temporal dimensions.

As for the timestep $t$, It utilizes the scalable adaptive layer normalization (S-AdaLN)~\cite{ma2024latte}. This linear layer will compute the $\gamma$, $\beta$, and $\alpha$ based on the timestep embedding $c$. After the layer normalization, the scale and shift operation is $\gamma_1LayerNorm(z)+\beta$, where the $h$ denotes the hidden embedding within the transformer blocks. Before the residual connections, the scale operation is $\alpha h$.

Suppose we have the latent feature $z \in \mathbb{R}^{B\times C\times T\times H\times W}$, where $B$ denotes the batch size, $C$ the channel count of video frames, $T$ the number of video frames, and $H$ and $W$ the height and width of each latent feature, respectively. The latent feature should be reshaped to the $z_s \in \mathbb{R}^{B_s\times S \times d}$ before the Spatial Self-Attention Block, where $B_s=B\times T$ and $S=H\times W$. Subsequently, this feature will be reshaped to the $z_t \in \mathbb{R}^{B_t\times T \times d}$ before the Temporal Self-Attention Block, where $B_t=B\times S$.

To introduce the LLM embedding, we introduce an LLM Cross-Attention Block immediately following the T5 Cross-Attention Block. This addition enhances the model's ability to incorporate contextual information from large language models, thereby enriching the feature representation. To regulate the influence of this cross-attention mechanism, a gating coefficient $\gamma$ is employed, initially set to zero, allowing for gradual integration and fine-tuning of the LLM-derived features.

\section{Training Prompts Recaption}
Incorporating detailed prompts into model training significantly enriches the realism and consistency of generated scenes. By providing explicit instructions regarding setting, actions, and other cinematic elements, these prompts help the model to generate more accurate complex environments and interactions. To advance the model's ability to process such detailed prompts, we employ a multimodal large language model (MLLM) for training in prompt re-captioning.

As shown in Fig.~\ref{supp_fig: recap1}, under the help of LLaVA-NEXT, we utilize the instruction to generate a desirable detailed prompt. Compared to the original prompts, it presents the key elements like the setting, characters, and their actions. Expand on these by adding specifics about clothing, gestures, and expressions. Describe the lighting, atmosphere, and any notable objects to establish the mood. Furthermore, it also specifies camera details, such as angle, movement, and shot type, to convey the visual presentation. Mention the video style, such as documentary, to provide storytelling context. Finally, highlight the character interactions to suggest the tone and nature of their conversation.

More specifically, for the subject, the original prompt only mentions that there are two individuals, with one person seated and the other standing, while the detailed prompt specifies not only the positions of the individuals but also includes intricate details about their clothing. Such details help establish character identity and contribute to the narrative context within the scene. Moreover, the detailed prompt also points out ighting conditions, noting that the light source is a single overhead light. This creates a focused, dramatic atmosphere, casting distinct shadows and highlighting certain aspects of the scene, which can influence the viewer's emotional response and interpretation.

\section{Detailed Scores of Main Results}
Table.~\ref{supp_tab: vb1} and Table.~\ref{supp_tab: vb2} show the detailed score of each dimension of the Standard VBench~\cite{huang2023vbench} Prompt evaluation in the main table.

\section{More Qualitative Results}
In the Fig.~\ref{supp_fig: vs1},~\ref{supp_fig: vs2},~\ref{supp_fig: vs3},~\ref{supp_fig: vs4}, we show more qualitative results compared with baselines.

\newpage
\clearpage
{
    \small
    \bibliographystyle{ieeenat_fullname}
    \bibliography{main}
}

%% file: supp_asset/figure_stack.tex
\begin{figure}[h]
  \centering
  \includegraphics[width=0.95\linewidth]{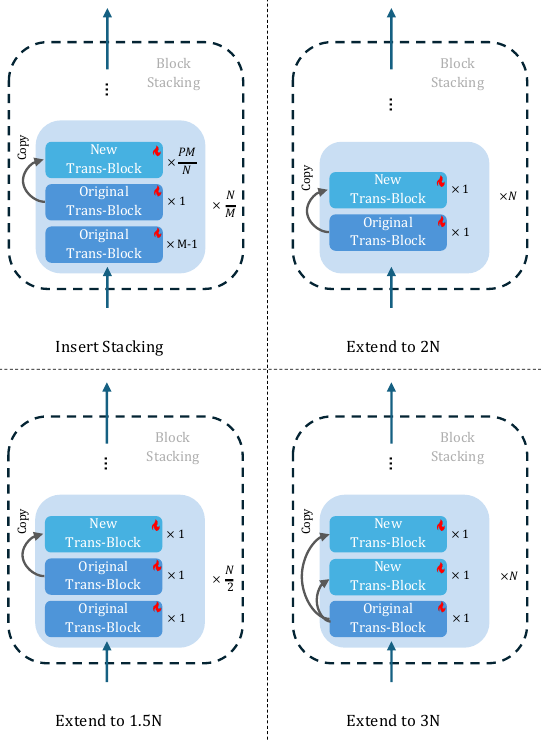}
  \caption{Details of Insert Stacking. We also show the three examples with extending to different numbers of transformer blocks.}
  \vspace{5mm}
  \label{supp_fig: stack}
\end{figure}

%% file: supp_asset/figure_arc.tex
\begin{figure*}[ht]
  \centering
  \includegraphics[width=0.95\linewidth]{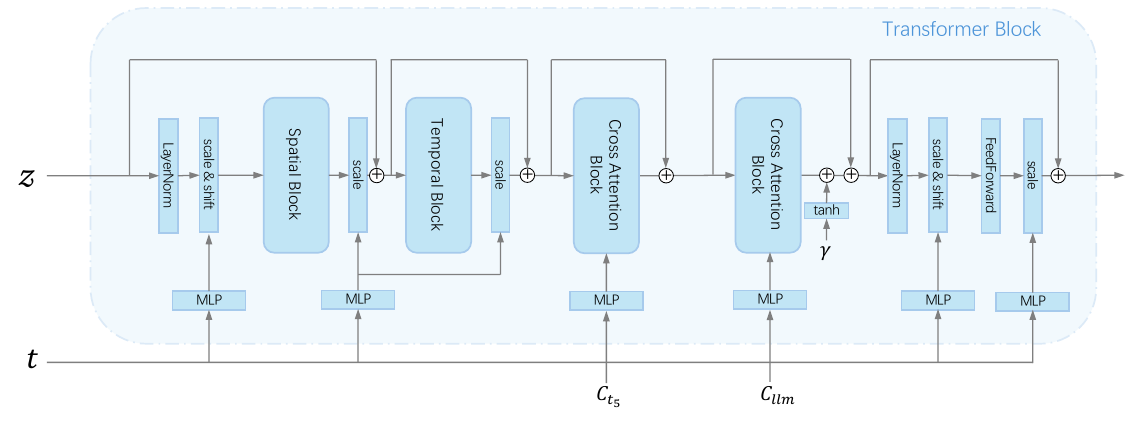}
  \caption{Overview of the transformer block of the text-to-video generation model. We modify the architecture of the transformer block by adding another cross-attention block, aiming to learn the condition of LLM text embedding. The LLMs cross-attention block follows the original T5 cross-attention block to enhance the language understanding ability of the generation models.}
  \vspace{5mm}
  \label{supp_fig: arc}
\end{figure*}

%% file: supp_asset/figure_recap1.tex
\begin{figure*}[ht]
  \centering
  \includegraphics[width=0.95\linewidth]{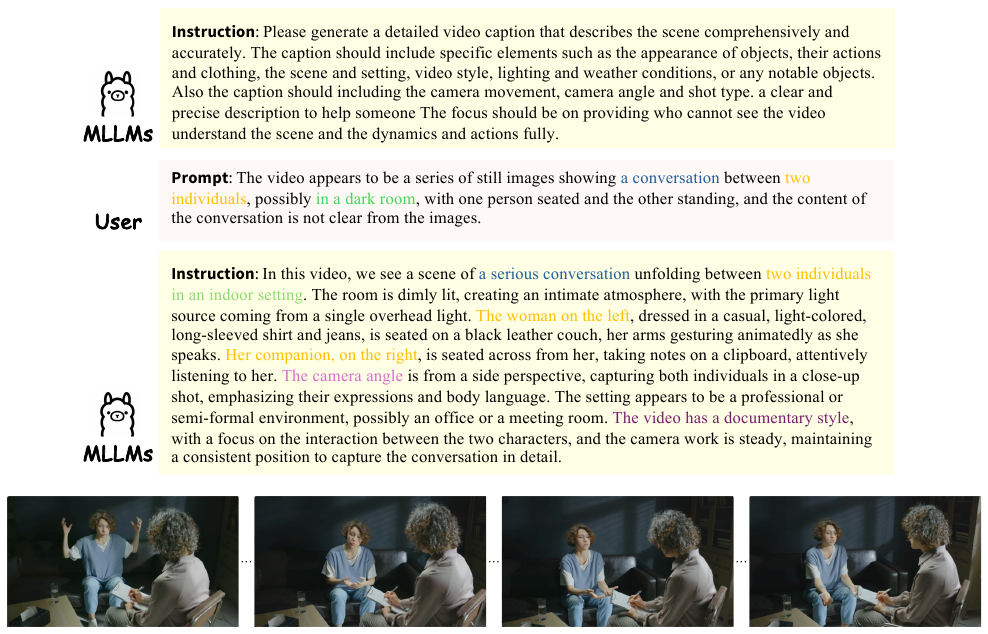}
  \caption{Example of training prompts recaption.}
  \vspace{5mm}
  \label{supp_fig: recap1}
\end{figure*}

%% file: supp_asset/table_vb1.tex
\begin{table*}[ht!]
\centering
\renewcommand{\arraystretch}{1.5}
\resizebox{\textwidth}{!}{
\setlength{\tabcolsep}{4pt} 
\begin{tabular}{lccccccc}
\toprule
\multicolumn{1}{c}{\multirow{3}{*}{\textbf{Method}}} & \multicolumn{7}{c}{\small{\textbf{VBench Standard Prompts on Quality Dimension}~\cite{huang2023vbench}}}                                             \\
                           & \makecell[c]{\small{Subject}\\ \small{Consistency}}$\uparrow$ & \makecell[c]{\small{Background} \\ \small{Consistency}}$\uparrow$ & \makecell[c]{\small{Temporal}\\ \small{Flickering}}$\uparrow$ & \makecell[c]{\small{Motion} \\ \small{Smoothness}}$\uparrow$ & \makecell[c]{\small{Aesthetic} \\ \small{Quality}}$\uparrow$ & \makecell[c]{\small{Imaging} \\ \small{Quality}}$\uparrow$ & \makecell[c]{\small{Dynamic} \\ \small{Degree}}$\uparrow$ \\ \midrule
Base model                 & 0.9159           & 0.9705 	& 0.9737 	& 0.9534 	& 0.5964 	& 0.6079 	& 0.5306    \\ 
Vanilla-0.7B                & 0.9069 	& 0.9779 	& 0.9820 	& 0.9474 	& 0.6029 	& 0.5850 	& 0.5778     \\ 
LoRA-0.7B~\cite{hu2021lora}   & 0.9343 	& 0.9768 	& 0.9805 	& 0.9603 	& 0.6154 	& 0.6124 	& 0.5028 \\
Expansion-1.4B             & 0.9212 	& 0.9807 	& 0.9837 	& 0.9583 	& 0.6139 	& 0.6225 	& 0.5833 \\ 
Expansion-1.4B-LLM         & 0.9637 	& 0.9843 	& 0.9872 	& 0.9761 	& 0.6413 	& 0.6431 	& 0.3056  \\
\bottomrule
\end{tabular}}
\caption{Quantitative results of the quality score on the VBench standard prompts.}
\label{supp_tab: vb1}
\end{table*}

\begin{table*}[ht!]
\centering
\resizebox{\textwidth}{!}{
\begingroup
\setlength{\tabcolsep}{4pt} 
\renewcommand{\arraystretch}{1.5} 
\begin{tabular}{lccccccccc}
\toprule
\multicolumn{1}{c}{\multirow{3}{*}{\textbf{Method}}}& \multicolumn{9}{c}{\small{\textbf{VBench Standard Prompts on Semantic Dimension}~\cite{huang2023vbench}}}                                             \\
                           & \makecell[c]{\small{Object} \\ \small{Class}}$\uparrow$ & \makecell[c]{\small{Multiple}\\ \small{Objects}}$\uparrow$ & \makecell[c]{\small{Human} \\ \small{Action}}$\uparrow$ & \makecell[c]{Color}$\uparrow$ & \makecell[c]{\small{Spatial}\\ \small{Relationship}}$\uparrow$ & \makecell[c]{Scene}$\uparrow$ & \makecell[c]{\small{Appearance} \\ \small{Style}}$\uparrow$ & \makecell[c]{\small{Temporal} \\ \small{Style}}$\uparrow$ & \makecell[c]{\small{Overall} \\ \small{Consistency}}$\uparrow$ \\ \midrule
Base model                  & 0.8612 	& 0.3107 	& 0.9240 	& 0.7774 	& 0.3268 	& 0.4372 	& 0.2439 	& 0.2505 	& 0.2764  \\ 
Vanilla-0.7B                & 0.8913 	& 0.3049 	& 0.8800 	& 0.7941 	& 0.2953 	& 0.4493 	& 0.2367 	& 0.2470 	& 0.2760  \\ 
LoRA-0.7B~\cite{hu2021lora} & 0.8763 	& 0.3163 	& 0.9120 	& 0.8447 	& 0.2942 	& 0.4578 	& 0.2406 	& 0.2501 	& 0.2747  \\
Expansion-1.4B              & 0.8541 	& 0.3438 	& 0.8780 	& 0.8360 	& 0.3330 	& 0.4647 	& 0.2393 	& 0.2533 	& 0.2821  \\ 
Expansion-1.4B-LLM          & 0.8922 	& 0.3229 	& 0.9500 	& 0.8131 	& 0.2466 	& 0.5408 	& 0.2378 	& 0.2533 	& 0.2720  \\
\bottomrule
\end{tabular}
\endgroup}
\caption{Quantitative results of the semantic score on the VBench standard prompts.}
\label{supp_tab: vb2}
\end{table*}

%% file: supp_asset/figure_vs.tex
\begin{figure*}[ht]
  \centering
  \includegraphics[width=0.82\linewidth]{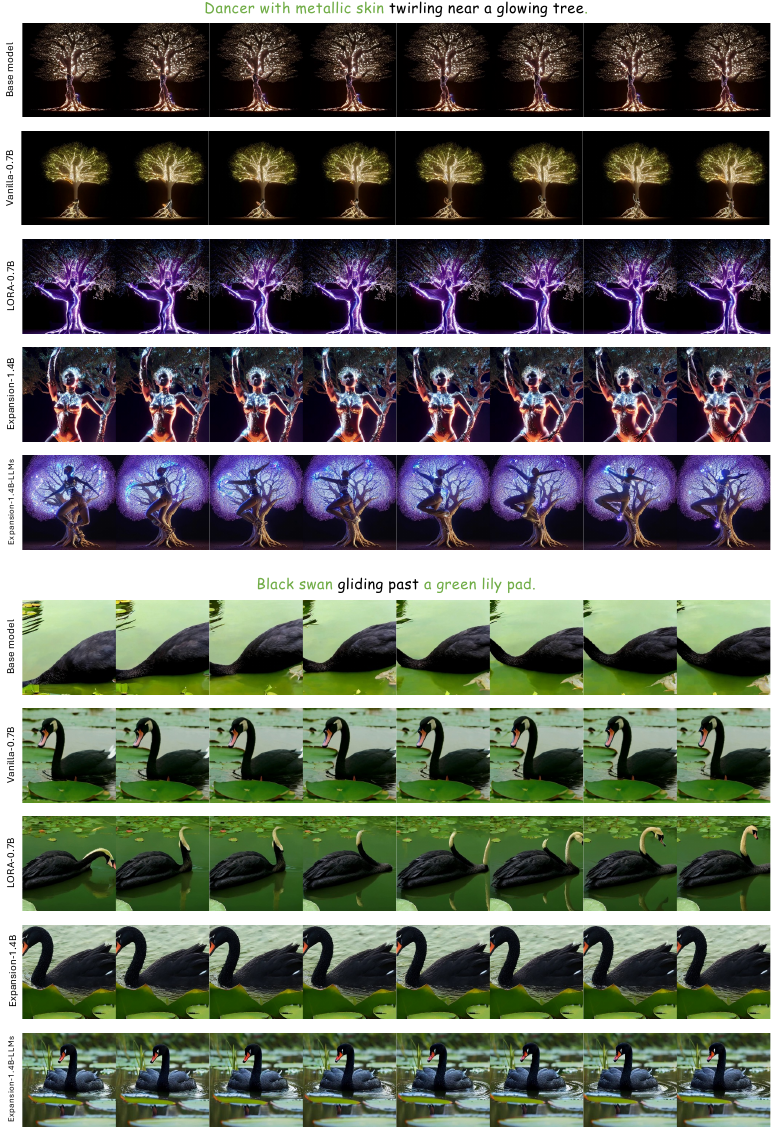}
  \caption{Qualitative results of our results compared with baselines.}
  \label{supp_fig: vs1}
\end{figure*}

\begin{figure*}[ht]
  \centering
  \includegraphics[width=0.82\linewidth]{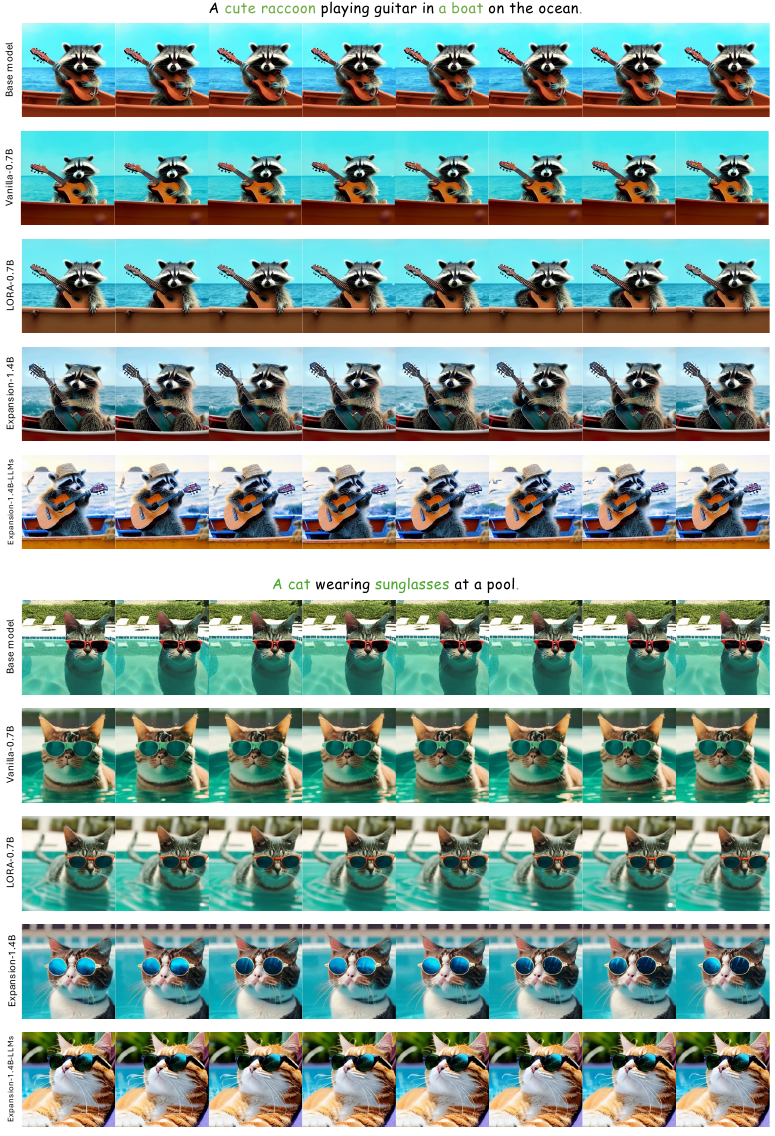}
  \caption{Qualitative results of our results compared with baselines.}
  \label{supp_fig: vs2}
\end{figure*}

\begin{figure*}[ht]
  \centering
  \includegraphics[width=0.82\linewidth]{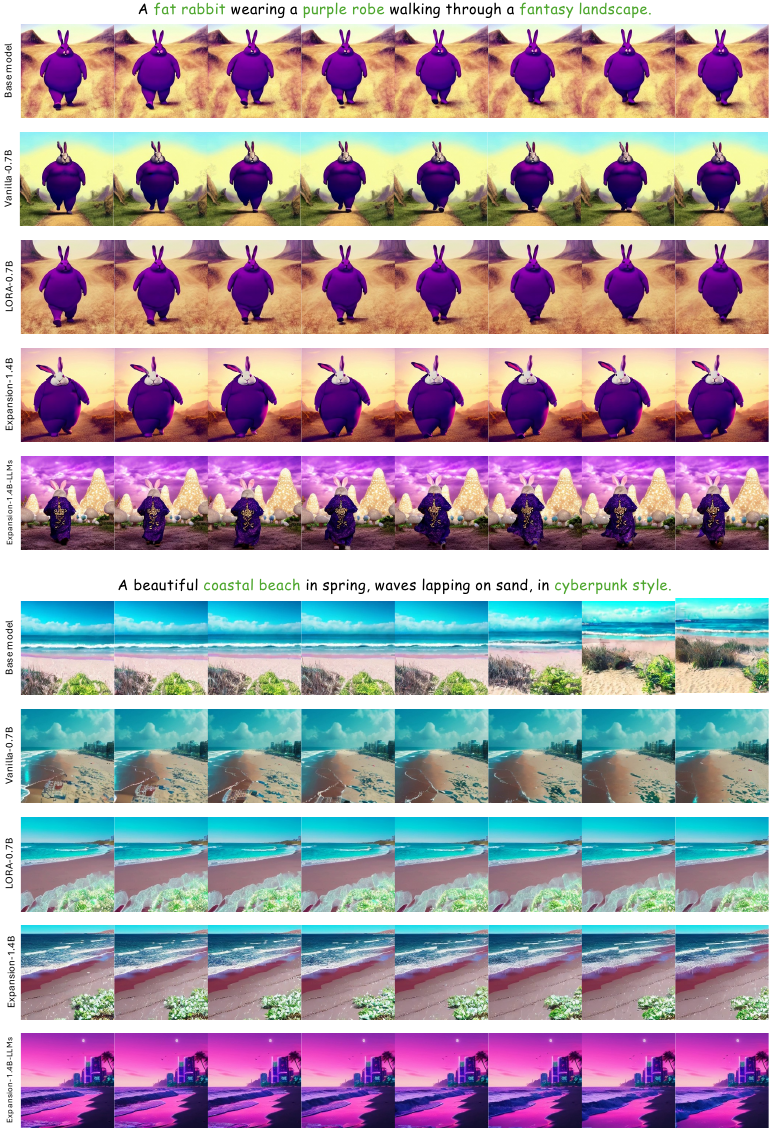}
  \caption{Qualitative results of our results compared with baselines.}
  \label{supp_fig: vs3}
\end{figure*}

\begin{figure*}[ht]
  \centering
  \includegraphics[width=0.82\linewidth]{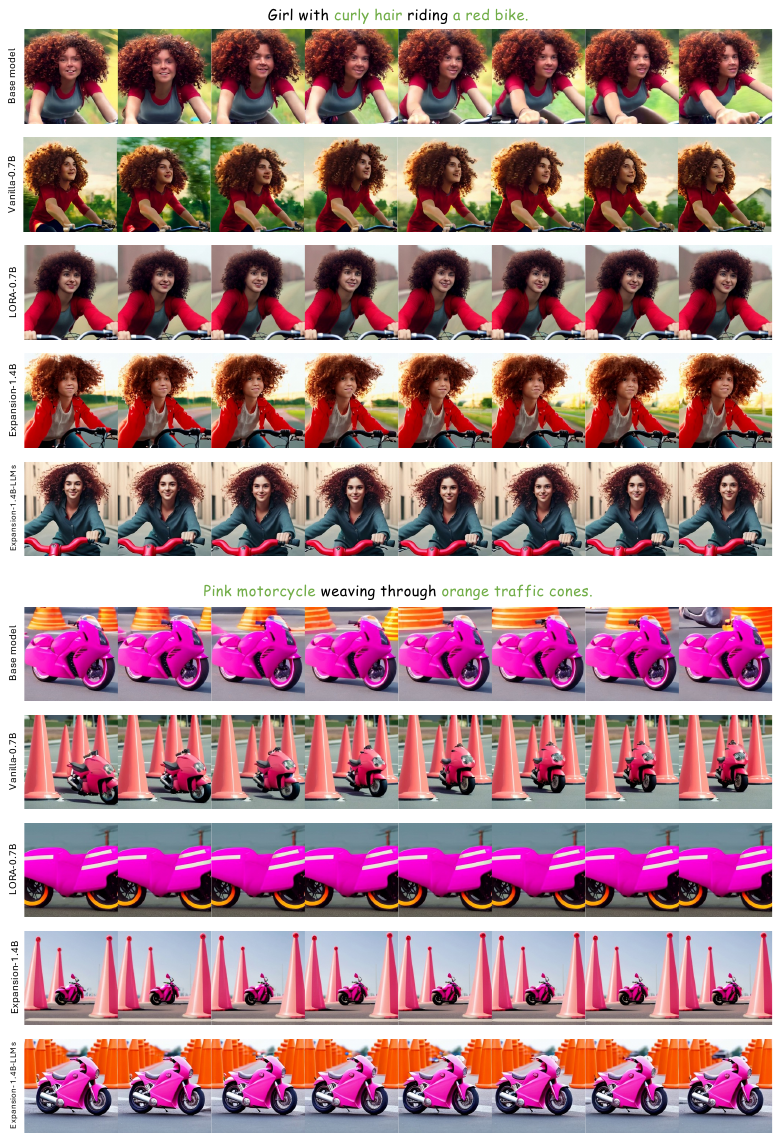}
  \caption{Qualitative results of our results compared with baselines.}
  \label{supp_fig: vs4}
\end{figure*}